\documentclass[]{spie}  

\usepackage{amsmath,amsfonts,amssymb}
\usepackage{graphicx}
\usepackage{nicematrix}
\usepackage{algorithm2e}
\usepackage{algpseudocode}
\usepackage{xcolor}
\usepackage{nicematrix}
\usepackage[colorlinks=true, allcolors=blue]{hyperref}
\usepackage{tikz}
\usepackage{tikzit}

\tikzstyle{black rectangle}=[tikzit fill=white, fill=white, draw=black, shape=rectangle, tikzit draw=black, align=center]
\tikzstyle{empty rectangle}=[fill=white, draw=white, shape=rectangle, tikzit fill=white, tikzit draw=white, align=center]

\tikzstyle{dashed line}=[-, dashed, fill=white, tikzit draw=black, tikzit fill=white, draw=black]

\title{Indecision Trees: Learning Argument-Based Reasoning under Quantified Uncertainty}

\author[a]{Jonathan S. Kent}
\author[b]{David H. Ménager Ph.D}
\affil[a]{Lockheed Martin Advanced Technology Center, Sunnyvale, California, USA}
\affil[b]{Parallax Advanced Research, Beavercreek, Ohio, USA}

\authorinfo{Send correspondence to Jonathan S. Kent: jonathan.s.kent@lmco.com. Work by Dr. Ménager was performed while at Lockheed Martin}

\begin{document} 
\maketitle

\begin{abstract}
Using Machine Learning systems in the real world can often be problematic, with inexplicable black-box models, the assumed certainty of imperfect measurements, or providing a single classification instead of a probability distribution.

This paper introduces Indecision Trees, a modification to Decision Trees which learn under uncertainty, can perform inference under uncertainty, provide a robust distribution over the possible labels, and can be disassembled into a set of logical arguments for use in other reasoning systems.  
\end{abstract}


\section{Introduction}

For many years, researchers and practitioners have designed and employed rule-based classification systems. In contrast with other machine learning classifiers, rule-based systems are useful, not only because of their predictive power, but also because they often produce knowledge structures, or rules, humans readily understand.
Such rules are typically learned from datasets which may contain a mixture of numeric and discrete features. Thus, there is a large variety of different types of rule-based classification systems. We cannot discuss each one in this paper, but we briefly desribe a few of them here. For example, MLEM2 \cite{grzymala2010local} discovers classification rules by finding local coverings of concepts blocks which may be approximated by rough sets. The AQ algorithm \cite{cervone2010algorithm} learns rules from a set of examples and counter examples.
The system tries to learn discriminative rules that cover the positive cases without covering the counter examples. Other rule learning systems like FURIA \cite{huhn2009furia} and RIPPER \cite{cohen1995learning} learn fuzzy rules. 
Another approach involves learning rules by inducing a decision tree structure. Several algorithms for this exist for building these trees, like CART \cite{breiman1984classification}, and ID3 \cite{quinlan1986induction}, but C4.5 \cite{quinlan1993program} is most commonly used.  
More recently, there have been rule-based systems which employ stochastic local search optimization techniques like simulated annealing, iterated local search, and ant colony optimization to learn rules \cite{jabbar2021rule, parpinelli2002data}.

The systems we described are impressive in their own right, but one area that deserves more attention in rule-based classification systems is handling quantified measurement noise in the data. More mainstream machine learning systems (i.e. Bayesian Networks and Neural Networks) have done a better job handling quantified uncertainty, but measurement noise associated with sensor readings still poses a significant challenge for rule induction algorithms.
Toward that end, we present a novel rule induction algorithm we call an \textit{indecision tree} capable of learning rules from quantified uncertainty from numeric and discrete-valued features. Our proposed solution is a decision tree that has been disassembled, reworking each step to operate under quantified uncertainty, and then reassembled. 
Thus, in order to understand indecision trees, we must first explain decision trees from the ground up.

The next section describes preliminaries of decision trees. Following this, we cover indecision trees in detail. Then, we present our experimental evaluations, discussing our design and results, before we conclude.

\section{Decision Trees}

\begin{figure}[t]
    \centering
    \small
    \begin{tabular}{r||c|c|c|c||l}
        idx & Height & Tail & Weight & Color & Breed  \\
        \hline
        1  & 23 & 10 & 76 & Blond & Retriever \\
        2  & 13 & 6 & 21 & Brown & Beagle \\
        3  & 24 & 10 & 39 & Black & Collie \\
        4  & 21 & 10 & 32 & Brown & Collie \\
        5  & 16 & 12 & 23 & Brown & Beagle \\
        6  & 19 & 11 & 65 & Brown & Retriever \\
        7  & 22 & 10 & 42 & Brown & Collie \\
        8  & 20 & 4 & 68 & Black & Retriever \\
        9  & 15 & 9 & 24 & Brown & Beagle \\
        10 & 22 & 7 & 72 & Blond & Retriever \\
    \end{tabular}
    \caption{An example dataset consisting of physical features of dogs, such as their shoulder height and tail length in inches, weight in pounds, and fur color, as well as a label representing the true breed of the dog.}
    \label{fig:dog_dataset}
\end{figure}

In order to explain the algorithm used to create decision trees, we will use the example of a dog breed classifier. 

Decision trees work by repeatedly, greedily splitting the data according to how much information is gained by doing so \cite{ross1993c4, breiman1984classification, hastie2009elements}. This information gain can be measured with a couple different metrics, but for our purposes, we will use entropy, from Information Theory \cite{shannon1948mathematical}. 

\subsection{Entropy}

Entropy, essentially, is a measure of how long an average description has to be to represent every element of a set of data. For example, if a dataset of dogs contained only retrievers, you could immediately identify the breed of a dog merely by knowing it belonged to the dataset, 0 extra information is needed. But if the dataset was 50\% retrievers and 50\% collies, you would need to be told which of the two possible breeds a given dog was, a single binary piece of information, beyond just knowing it belonged to the dataset. Thus, the entropy of the dataset is 1 bit.

However, if the dataset was, say, 70\% retrievers and 30\% collies, you have more information than in the 50:50 case, because you know that a given dog belonging to the dataset was more likely to be a retriever than a collie. The full expression for the entropy of a dataset $\mathcal{D}$ captures this, by defining entropy as the proportion of the dataset belonging to each class $c_i$, times how many bits would be needed to split the dataset into equal subsets the size of that class, summed over all the classes present:

$$H(\mathcal{D}) = \sum_{c_i \in \mathcal{D}} p(c_i) \log_2\Big(\frac{1}{p(c_i)}\Big)$$

which can be simplified as

$$H(\mathcal{D}) = -\sum_{c_i \in \mathcal{D}} p(c_i) \log_2\big(p(c_i)\big)$$

for computational purposes. For example, in the case of a 70:30 split as described, this produces an entropy of approximately 0.881

\subsection{Information Gain via Splitting with Clauses}

Let's start looking at the dataset in Figure \ref{fig:dog_dataset}. This dataset is composed of 40\% retrievers, 30\% beagles, and 30\% collies, for an entropy of 1.571. This is how much information we need to correctly label every element of the dataset. But the features included can tell us this information, and we can make use of them by asking questions about the features. For example, consider asking of each dog ``Is this dog 20 inches or taller at the shoulder?", and then sorting the dogs based on the answer. Of the dogs for which the answer is ``Yes," there are 3 retrievers and 3 collies, and of the dogs for which the answer is ``No," there are 3 beagles and 1 retriever.

We can rephrase this question into a clause, $C_{\textrm{height} \ge 20}(x_i) = $``The dog represented by $x_i$ is 20 inches or taller at the shoulder," and then divide the dogs based on whether $C$ is true or false, i.e.

\begin{center}
    \begin{tabular}{ccccccc}
         $\mathcal{D}_{C}$ & $=$ & $\{$ & $x_i \in \mathcal{D}$ & $|$ & $C(x_i)$ & $\}$\\
         $\mathcal{D}_{\neg C}$ & $=$ & $\{$ & $x_i \in \mathcal{D}$ & $|$ & $\neg C(x_i)$ & $\}$\\
    \end{tabular}
\end{center}

and then look at the entropy of the split data, getting $H(\mathcal{D}_{C_{\textrm{height} \ge 20}}) = 1$ and $H(\mathcal{D}_{\neg C_{\textrm{height} \ge 20}}) = 0.811$. By taking the weighted average of the entropy, we get, on average, how much entropy remains in the dataset once we've applied this clause, and by subtracting this entropy from the original, we get the information gain of the clause itself;

$$IG(\mathcal{D}, C) = H(\mathcal{D}) - \frac{|\mathcal{D}_{C}| \cdot H(\mathcal{D}_{C}) + |\mathcal{D}_{\neg C}| \cdot H(\mathcal{D}_{\neg C})}{|\mathcal{D}|}$$

which in this case is 0.646, the number of bits of information that the clause $C_{\textrm{height} \ge 20}$ gives us.

\subsection{Finding the Best Clause}

Stating the best clause is as simple as

$$C_*(\mathcal{D}) = \arg \max_{C \in \mathcal{C}} IG(\mathcal{D}, C)$$

essentially, enumerating all possible clauses as $\mathcal{C}$, and then trying all of them to see which one has the greatest information gain. The question is how to enumerate clauses.

\subsubsection{Discrete Feature Clauses}

Each clause of a discrete feature is defined by some subset of the discrete feature values. For example, consider the possible fur colors of the dogs: black, blond, and brown. Each clause would take a form like $C_{\textrm{fur} \in \textrm{\{black\}}}(x_i) = $ ``The dog represented by $x_i$ has black fur" or $C_{\textrm{fur} \in \textrm{\{black, blond\}}}(x_i) = $ ``The dog represented by $x_i$ has black or blond fur." For the $j^\textrm{th}$ discrete feature $f^j_d$, all the possible clauses are enumerated with

$$\mathcal{C}^j_d = \{C_{f^j_d \in s} | s \in \mathcal{P}(f^j_d) \}$$

but in practice, half of these clauses are ignored, as each clause has an opposite clause that results in the same split, just backwards, e.g. ``The dog represented by $x_i$ is blond" results in the same split as ``The dog represented by $x_i$ is brown or black." Once that's done, all the possible discrete clauses can be collected together,

$$\mathcal{C}_d = \bigcup_j \mathcal{C}^j_d$$

then it's just a matter of enumerating the continuous clauses.

\subsubsection{Continuous Feature Clauses}

The enumeration of the discrete features relies on the power set of the discrete features being finite, which requires that the set of discrete features is finite. But continuous features are assumed to have an infinite number of feature values, and furthermore that nearby feature values should typically be grouped together anyway. Thus, instead of futilely trying to enumerate every possible continuous clause, we will instead enumerate clauses that can be represented by threshold values, such as $C_{\textrm{height} \ge 20}$, which was discussed earlier. 

For the $j^\textrm{th}$ continuous feature $f^j_c$, while there are infinite possible threshold values, most thresholds are either useless - being above or below the maximum or minimum values, respectively - known suboptimal - dividing two sequential elements which are from the same class - or equivalent to other thresholds - making the same split in the training data as another. So the set of thresholds we bother checking, $T^j$, is simply the set of middles between any two values of $f^j_d$, provided that those two values are sequential if you sorted the data based on that feature, and that they belong to samples of different classes. For example, looking at heights: the values 27 and 10 wouldn't be checked as thresholds because they're out of range, 15.5 wouldn't be checked because it separates two beagles, and 17 wouldn't be checked because it would result in the same split as 17.5. The threshold values that would be checked are 17.5, 20.5, 21.5, and 23.5. So the clauses for the feature $f^j_c$ would be enumerated with

$$\mathcal{C}^j_c = \{C_{f^j_c \ge t} | t \in T^j\}$$

and then collected over all the features with

$$\mathcal{C}_c = \bigcup_j \mathcal{C}^j_c$$

at which point all the clauses to be checked have now been enumerated, and we put the continuous and the discrete feature clause collections together with

$$\mathcal{C} = \mathcal{C}_d \cup \mathcal{C}_c$$

meaning that $C_*$ can now be found by iterating through and checking all the information gains.

\subsection{Growing the Decision Tree}

Starting from the root node, the decision tree is grown recursively by associating each node with the best clause to be found at that node based on the current state of the dataset, and then passing the dataset subject to both that clause and its negation to that node's left and right children, respectively. By holding onto some information regarding the state of the tree over time, such as the current entropy of the subset of the dataset, facts about the current best clause, or simply the current depth in the tree, the algorithm can produce a leaf instead of another branch. This means simply writing down the probability of each class as they currently stand, and ceasing to branch. The algorithm for this is shown in Algorithm \ref{alg:grow_dec_tree}. 

{\centering
\begin{minipage}[t]{0.6\linewidth}
\begin{algorithm}[H]
\SetAlgoLined
\DontPrintSemicolon

    \SetKwFunction{FMain}{Node}
    \SetKwFunction{FUpdate}{Update}
    \SetKwFunction{FBreak}{Should\_Leaf}

    \KwIn{$\mathcal{D}, b$, \FBreak, \FUpdate}
    \KwOut{$\mathcal{N}$ \Comment{Root Node}}

    \SetKwProg{Fn}{Function}{:}{}
    \Fn{\FMain{$\mathcal{D}_q, b$}}
    {
        {
        \eIf {\FBreak{$b$}}
        {
            {$\mathcal{N}_q.leaf \longleftarrow \textrm{true}$}\;
            {$\mathcal{N}_q.p \longleftarrow \{\big(c_i, p(c_i)\big) | c_i \in \mathcal{D}_q$\}}\;
        }
        {
            {$\mathcal{N}_q.leaf \longleftarrow \textrm{false}$}\;
            {$\mathcal{N}_q.C \longleftarrow C_*(\mathcal{D}_q)$}\;
            {$b \longleftarrow$ \FUpdate{$b, \mathcal{D}_q, \mathcal{N}_q.C$}}\;
            {$\mathcal{N}_q.left\_child \longleftarrow$ \FMain{$\mathcal{D}_{q \cup \{C\}}, b$}}\;
            {$\mathcal{N}_q.right\_child \longleftarrow$ \FMain{$\mathcal{D}_{q \cup \{\neg C\}}\; b$}}\;
        }
        }
        
        \textbf{return}($\mathcal{N}_q$)\;
    }
    
    {$q \longleftarrow \emptyset$}\;
    {$\mathcal{D}_q \longleftarrow \mathcal{D}$}\;
    {$\mathcal{N} \longleftarrow$ \FMain{$\mathcal{D}_q, b$}}\;

\caption{Growing a Decision Tree}
\label{alg:grow_dec_tree}
\end{algorithm}
\end{minipage}
\par
}

\subsection{Using the Decision Tree}

Once this is done, inferences on new samples can be pulled out using Algorithm \ref{alg:inf_dec_tree}. All this does is, starting at the root node, go to each child according to whether or not the sample satisfies the clause at that node. Once it hits a leaf node, it has found the probabilities that the tree outputs. To use this for classification, simply choose the class with the highest associated probability.

{\centering
\begin{minipage}[b]{0.6\linewidth}
\begin{algorithm}[H]
\SetAlgoLined
\DontPrintSemicolon

    \SetKwFunction{FMain}{Inference}

    \KwIn{$x_k, \mathcal{N}$}
    \KwOut{$P = \{\big(c_i, p(x_k \in c_i)\big)\}$ \Comment{Class Probabilities}}

    \SetKwProg{Fn}{Function}{:}{}
    \Fn{\FMain{$x_k, \mathcal{N}_q$}}
    {
        {
        \eIf {$\mathcal{N}_q$.leaf}
        {
            \textbf{return}($\mathcal{N}_q.p$)\;
        }
        {
        {
            \eIf {$\mathcal{N}_q.C(x_k)$}
            {
                \textbf{return}(\FMain{$x_k, \mathcal{N}_q.$\textit{left\_child}})\;
            }
            {
                \textbf{return}(\FMain{$x_k, \mathcal{N}_q.$\textit{right\_child}})\;
            }
            }
        }
        }
    }
    
    {$q \longleftarrow \emptyset$}\;
    {$\mathcal{N}_q \longleftarrow \mathcal{N}$}\;
    {$P \longleftarrow$ \FMain{$x_k, \mathcal{N}_q$}}\;

\caption{Inference Using a Decision Tree}
\label{alg:inf_dec_tree}
\end{algorithm}
\end{minipage}
\par
}

It also becomes possible to turn the decision tree into a set of logical rules. This is because, for each leaf in the tree, there is exactly one way for samples to reach it. If a leaf is represented by node $\mathcal{N}_q$, where $q$ is a set of clauses $\{C_1, C_2, C_3, \dots, C_n\}$, then a rule can take the form of ``Satisfying all clauses in $q \implies$ inferred class probabilities are $\mathcal{N}_q.p$," with the added note that clauses can be condensed according to their redundancy. For example, if one clause takes the form ``Fur length is less than 10," and another is ``Fur length is less than 15," the latter can simply be thrown out. Similarly would be clauses like ``Fur color is brown," and ``Fur color is blond or brown." This process is shown in Algorithm \ref{alg:rul_dec_tree}.

{\centering
\begin{minipage}[b]{0.6\linewidth}
\begin{algorithm}[H]
\SetAlgoLined
\DontPrintSemicolon

    \SetKwFunction{FMain}{Get\_Rules}
    \SetKwFunction{FCondense}{Condense}

    \KwIn{$\mathcal{N}$, \FCondense}
    \KwOut{$R = \{r_i\}$ \Comment{Set of Rules}}

    \SetKwProg{Fn}{Function}{:}{}
    \Fn{\FMain{$\mathcal{N}_q$}}
    {
        {
        \eIf {$\mathcal{N}_q$.leaf}
        {
            \textbf{return}($\{($\FCondense$(q), \mathcal{N}_q.p)\}$)\;
        }
        {
            \textbf{return}(\FMain{$\mathcal{N}_q.$\textit{left\_child}} $\cup$ \FMain{$\mathcal{N}_q.$\textit{right\_child}})\;
        }
        }
    }
    
    {$q \longleftarrow \emptyset$}\;
    {$\mathcal{N}_q \longleftarrow \mathcal{N}$}\;
    {$R \longleftarrow$ \FMain{$x_k, \mathcal{N}_q$}}\;

\caption{Pulling Rules from a Decision Tree}
\label{alg:rul_dec_tree}
\end{algorithm}
\end{minipage}
\par
}

With this established, we will now begin to explain indecision trees: how they operate, their similarities to decision trees, and the modifications required to grow and infer from them under uncertainty.

\section{Indecision Trees}

Indecision trees are a modified version of decision trees, to accommodate quantified measurement uncertainty. Fundamentally, this works by considering each subset of the data, not to actually contain any given data point, but instead for each data point to have some probability of existing in each subset. For example, if a dog (given by idx 10 in Figure \ref{fig:unc_dog_dataset}) is observed to have a height of 22 inches, with an uncertainty of 2.5 inches representing the standard deviation of the measurement, then rather than saying the dog is definitely shorter than 24 inches, we would say that the dog has a 79\% chance of being shorter than 24 inches, because $p\big(h \sim \mathcal{N}(\mu = 22, \sigma = 2.5) < 24\big) \approx 0.79$.

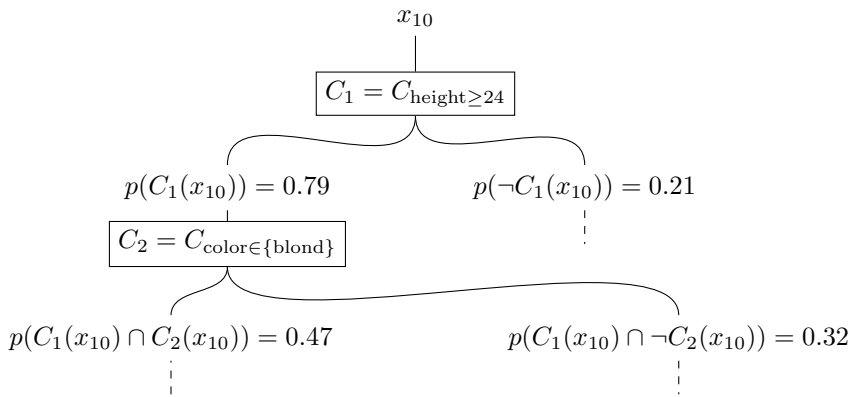
\begin{figure}[b]
{\centering
\begin{minipage}[b]{0.6\linewidth}
    \centering
    \begin{tikzpicture}
	\begin{pgfonlayer}{nodelayer}
		\node [style=black rectangle] (1) at (0, -1) {$C_1 = C_{\textrm{height} \geq 24}$};
		\node [style=empty rectangle] (2) at (0, 0) {$x_{10}$};
		\node [style=empty rectangle] (7) at (-2.5, -2.25) {$p(C_1(x_{10})) = 0.79$};
		\node [style=black rectangle] (8) at (-2.5, -3) {$C_2 = C_{\textrm{color} \in \{\textrm{blond}\}}$};
		\node [style=empty rectangle] (9) at (2.25, -2.25) {$p(\neg C_1(x_{10})) = 0.21 $};
		\node [style=empty rectangle] (10) at (-3.25, -4.25) {$p(C_1(x_{10}) \cap C_2(x_{10})) = 0.47$};
		\node [style=empty rectangle] (11) at (3.5, -4.25) {$p(C_1(x_{10}) \cap \neg C_2(x_{10})) = 0.32$};
		\node [style=none] (12) at (2.25, -3) {};
		\node [style=none] (14) at (-3.25, -5) {};
		\node [style=none] (15) at (3.5, -5) {};
	\end{pgfonlayer}
	\begin{pgfonlayer}{edgelayer}
		\draw (2) to (1);
		\draw [in=90, out=-90] (1) to (7);
		\draw (7) to (8);
		\draw [in=90, out=-90] (1) to (9);
		\draw [in=-90, out=90, looseness=0.50] (11) to (8);
		\draw [in=90, out=-90] (8) to (10);
		\draw [style=dashed line] (9) to (12.center);
		\draw [style=dashed line] (11) to (15.center);
		\draw [style=dashed line] (14.center) to (10);
	\end{pgfonlayer}
\end{tikzpicture}
\caption{Partial Indecision Tree with two clauses, omitting leaf nodes.}
\label{fig:ind_tree}
\end{minipage}
\par
}
\end{figure}

As a result, if that sample were subjected to the clause ``height is greater than 24 inches," rather than following one branch of the tree or the other, both branches are following simultaneously, according to the probabilities of the clause being satisfied or unsatisfied, and so on in turn. For example, the 79\% associated with the dog being shorter than 24 inches might go to another branch, with a clause like ``the dog is blond," and at that branch the clause might be satisfied 60\% of the time and unsatisfied 40\% of the time. By assuming that the features are independent, this would split the 79\% into 47\% and 32\%, respectively, would then each go to the next branches. This is shown in Figure \ref{fig:ind_tree}. At the end, the sum of probabilities over all of the leaves will necessarily be 100\%, and by summing the probability of reaching each leaf times the probability of being a given class once at that leaf, you receive the probability of the sample belonging to that class. With this as the formulation, we can begin implementing the modifications that turn a decision tree into an indecision tree.

\begin{figure*}[t]
    \centering
    \small
    \begin{NiceTabular}{r||cc|cc|cc|ccc||l}
        idx & \multicolumn{2}{c}{Height} & \multicolumn{2}{c}{Tail} & \multicolumn{2}{c}{Weight} & \multicolumn{3}{c}{Color} & Breed  \\
        & $\mu$ & $\sigma$ & $\mu$ & $\sigma$ & $\mu$ & $\sigma$ & $p(\textrm{Blond})$ & $p(\textrm{Brown})$ & $p(\textrm{Black})$ & \\
        \hline
        1  & 23 & 3 & 10 & 1 & 76 & 7 & 0.9 & 0.1 & 0 & Retriever \\
        2  & 13 & 1.5 & 6 & 0.5 & 21 & 3 & 0.2 & 0.7 & 0.1 & Beagle \\
        3  & 24 & 2.5 & 10 & 0.5 & 39 & 4 & 0 & 0.4 & 0.6 & Collie \\
        4  & 21 & 2 & 10 & 1.5 & 32 & 4 & 0 & 0.7 & 0.3 & Collie \\
        5  & 16 & 1.5 & 12 & 1 & 23 & 2 & 0.1 & 0.8 & 0.1 & Beagle \\
        6  & 19 & 1 & 11 & 2 & 65 & 5 & 0.1 & 0.5 & 0.4 & Retriever \\
        7  & 22 & 2 & 10 & 0.5 & 42 & 6 & 0.5 & 0.3 & 0.2 & Collie \\
        8  & 20 & 2.5 & 4 & 0.5 & 68 & 6 & 0.2 & 0.2 & 0.6 & Retriever \\
        9  & 15 & 1.5 & 9 & 1 & 24 & 2 & 0.3 & 0.4 & 0.3 & Beagle \\
        10 & 22 & 2.5 & 7 & 0.5 & 72 & 8 & 0.6 & 0.1 & 0.3 & Retriever \\
    \end{NiceTabular}
    \caption{An example of an uncertain dataset, in which each feature value is replaced with a distribution over its possible values. Rather than being given the exact height of each dog, for example, we are given the mean and standard deviation of the height of the dog, representing the measured value and the quantified measurement uncertainty. For the fur colors, we are given the probability of it being each color, representing a discrete distribution rather than a discrete feature value.}
    \label{fig:unc_dog_dataset}
\end{figure*}

\subsection{Entropy under Uncertainty}

In order to calculate entropy under these conditions, we need to rework how we approach the probability of each class appearing in an uncertain dataset $\mathcal{D}^u$. We cannot calculate $p(c_i)$ as being $p(x_k \in c_i | x_k \in \mathcal{D}^u)$ because it's not as simple as ``$x_k \in \mathcal{D}^u$" anymore. Instead, we have to use a weighted sum, with weights representing the probability of each sample appearing in the uncertain dataset, with

$$
p(c_i) = \frac{\sum_{x_l \in c_i} p(x_l \in \mathcal{D}^u)}{\sum_{x_k} p(x_k \in \mathcal{D}^u)}
$$

at which point the new $p(c_i)$ can be plugged into the original entropy formula verbatim, using  

$$H(\mathcal{D}^u) = -\sum_{c_i \in \mathcal{D}^u} p(c_i) \log_2\big(p(c_i)\big)$$

\subsection{Information Gain under Uncertainty}

For this, we can look at the uncertain dataset in Figure \ref{fig:unc_dog_dataset}. Similarly to the dataset in Figure \ref{fig:dog_dataset}, it is composed of 40\% retrievers, 30\% beagles, and 30\% collies. Because no clauses have yet been applied to the dataset, the probability of each sample appearing in the dataset as it exists is 100\%, so it has the same entropy as the certain dataset, 1.571.

We can then consider the same clause that was used in the certain case, $C_{\textrm{height} \ge 20}$, and then produce two subsets associated with this clause, 

\begin{center}
    \begin{tabular}{ccccc}
         $\mathcal{D}^u_{C}$ & $|$ & $p(x_i \in \mathcal{D}^u_{C}) $ & $ = $ & $  p\big(C(x_i)\big) \cdot p(x_i \in \mathcal{D}^u)$\\
          $\mathcal{D}^u_{\neg C}$ & $|$ & $p(x_i \in \mathcal{D}^u_{\neg C}) $ & $ = $ & $  p\big(\neg C(x_i)\big) \cdot p(x_i \in \mathcal{D}^u)$\\
    \end{tabular}
\end{center}

which gives the following probabilities:

\begin{center}
    \begin{tabular}{r||c|c||l}
        idx & $p(x_i \in \mathcal{D}^u_{C})$ & $p(x_i \in \mathcal{D}^u_{\neg C})$ & Breed\\
        \hline
        1  & 0.84 & 0.16 & Retriever\\
        2  & 0 & 1 & Beagle\\
        3  & 0.95 & 0.05 & Collie\\
        4  & 0.69 & 0.31 & Collie\\
        5  & 0 & 1 & Beagle\\
        6  & 0.16 & 0.84 & Retriever\\
        7  & 0.84 & 0.16 & Collie\\
        8  & 0.5 & 0.5 & Retriever\\
        9  & 0 & 1 & Beagle\\
        10 & 0.79 & 0.21 & Retriever\\
    \end{tabular}
\end{center}

allowing us to calculate the entropies for each dataset. Probabilistically, $\mathcal{D}^u_{C_{\textrm{height} \ge 20}}$ contains approximately 2.29 retrievers, 0 beagles, and 2.48 collies, with an entropy of approximately 0.99. Similarly, $\mathcal{D}^u_{\neg C_{\textrm{height} \ge 20}}$ is expected to contain 1.71 retrievers, 3 beagles, and 0.52 collies, with an entropy of approximately 1.32. While these are relatively close to the numbers in the discrete case, they are different enough that failing to account for the uncertainty would yield a quite different result. We can then modify the notion of cardinality by using the number of samples the dataset probably contains,

$$|\mathcal{D}^u_C| = \sum_{x_i}p(x_i \in \mathcal{D}^u_C)$$

to redo the information gain calculation

$$IG(\mathcal{D}^u, C) = H(\mathcal{D}^u) - \frac{|\mathcal{D}^u_{C}| \cdot H(\mathcal{D}^u_{C}) + |\mathcal{D}^u_{\neg C}| \cdot H(\mathcal{D}^u_{\neg C})}{|\mathcal{D}^u|}$$

which here takes a value of 0.405. This is the information gained from applying the clause $C_{\textrm{height} \ge 20}$ to the uncertain version of the dataset.

\subsection{Best Clause under Uncertainty}

The exact same procedure can be used to enumerate clauses for indecision trees as is used for decision trees. However, this is with one caveat; the information gain of continuous clauses is a piecewise constant function of the threshold value for decision trees, while it is continuous and differentiable with respect to the threshold value for indecision trees. As a result, simply guessing and checking will yield suboptimal results compared to a method of gradient descent starting from a variety of positions. However, this procedure would be computationally expensive when repeated ad nauseum in the enumeration of all possible clauses, and so it will be left to future research.

\subsection{Growing the Indecision Tree}

The process of growing an indecision tree is the same as growing a decision tree, with a switch from a certain to an uncertain dataset. The necessary modifications are shown in Algorithm \ref{alg:grow_ind_tree}.

{\centering
\begin{minipage}{0.6\linewidth}
\begin{algorithm}[H]
\SetAlgoLined
\DontPrintSemicolon

    \SetKwFunction{FMain}{Node}
    \SetKwFunction{FUpdate}{Update}
    \SetKwFunction{FBreak}{Should\_Leaf}

    \KwIn{$\mathcal{D}^u, b$, \FBreak, \FUpdate}
    \KwOut{$\mathcal{N}$ \Comment{Root Node}}

    \SetKwProg{Fn}{Function}{:}{}
    \Fn{\FMain{$\mathcal{D}^u_q, b$}}
    {
        {
        \eIf {\FBreak{$b$}}
        {
            {$\mathcal{N}_q.leaf \longleftarrow \textrm{true}$}\;
            {$\mathcal{N}_q.p \longleftarrow \{\big(c_i, p(c_i)\big) | c_i \in \mathcal{D}^u_q$\}}\;
        }
        {
            {$\mathcal{N}_q.leaf \longleftarrow \textrm{false}$}\;
            {$\mathcal{N}_q.C \longleftarrow C_*(\mathcal{D}^u_q)$}\;
            {$b \longleftarrow$ \FUpdate{$b, \mathcal{D}^u_q, \mathcal{N}_q.C$}}\;
            {$\mathcal{N}_q.left\_child \longleftarrow$ \FMain{$\mathcal{D}^u_{q \cup \{C\}}, b$}}\;
            {$\mathcal{N}_q.right\_child \longleftarrow$ \FMain{$\mathcal{D}^u_{q \cup \{\neg C\}}\; b$}}\;
        }
        }
        
        \textbf{return}($\mathcal{N}_q$)\;
    }
    
    {$q \longleftarrow \emptyset$}\;
    {$\mathcal{D}^u_q \longleftarrow \mathcal{D}^u$}\;
    {$\mathcal{N} \longleftarrow$ \FMain{$\mathcal{D}^u_q, b$}}\;

\caption{Growing an Indecision Tree}
\label{alg:grow_ind_tree}
\end{algorithm}
\end{minipage}
\par
}

\subsection{Using the Indecision Tree}

While pulling the rules out of the indecision tree remains exactly the same as pulling the rules out of a decision tree, performing inference with indecision tree does require some modification in process. Rather than simply finding the leaf node that contains the sample on which inference is being formed, every single leaf node has to be checked, the probabilities of reaching those leaf nodes multiplied by their class probabilities, and then returning the sum over all of those values. The algorithm with these modifications in place can be seen in Algorithm \ref{alg:inf_ind_tree}.

{\centering
\begin{minipage}{0.6\linewidth}
\begin{algorithm}[H]
\SetAlgoLined
\DontPrintSemicolon

    \SetKwFunction{FMain}{Inference}

    \KwIn{$x_k, \mathcal{N}$}
    \KwOut{$P = \{\big(c_i, p(x_k \in c_i)\big)\}$ \Comment{Class Probabilities}}

    \SetKwProg{Fn}{Function}{:}{}
    \Fn{\FMain{$x_k, \mathcal{N}_q$}}
    {
        {
        \eIf {$\mathcal{N}_q$.leaf}
        {
            \textbf{return}($\{(c_i, \mathcal{N}_q.p(c_i) \cdot \prod_{C \in q}p\big(C(x_k)\big))\}$)\;
        }
        {
            $P_l \longleftarrow$ \FMain{$x_k, \mathcal{N}_q.$\textit{left\_child}}\;
            $P_r \longleftarrow$ \FMain{$x_k, \mathcal{N}_q.$\textit{right\_child}}\;
            \textbf{return}($\{(c_i, P_l(c_i) + P_r(c_i))\}$)\;
        }
        }
    }
    
    {$q \longleftarrow \emptyset$}\;
    {$\mathcal{N}_q \longleftarrow \mathcal{N}$}\;
    {$P \longleftarrow$ \FMain{$x_k, \mathcal{N}_q$}}\;

\caption{Inference Using an Indecision Tree}
\label{alg:inf_ind_tree}
\end{algorithm}
\end{minipage}
\par
}

\section{Experimentation}

\subsection{Dataset}

Publicly available datasets that include quantified uncertainty of the form mentioned in this paper are rare, and so for the purposes of experimentation, a synthetic dataset was created. This dataset consists of thirteen features, with a three-way classification problem. Of the thirteen features, nine are continuous, and four are discrete.

Of the nine continuous features, five were generated via the following procedure: for each of the three classes, three five-dimensional centroid vectors were randomly generated, for a total of nine centroids. When a sample is selected to be from a given class, one of the centroid vectors associated with the class is selected, and perturbed with random noise, the standard deviations of which were also randomly selected, as well as those deviations differing by feature. These features are then recorded using their observed values as $\mu$, and their randomly pre-selected standard deviations as $\sigma$. This produces behavior where features correlate based on classes, which themselves cannot be simply represented as clusters. Of the remaining four continuous features, two are simply copies of previous features, one is the sum of two previous features, and one was generated without regard to the class, thus being a source of pure noise.

Of the four discrete features, two were generated by the following procedure: for each class, a probability was selected for each bin in a given feature, differing by class. Ground truth bins for each sample were sampled according to these probabilities. Each bin was then associated with a randomly generated centroid vector in a five dimensional vector space, and probabilities per bin were assigned according to the softmax of the negative distances to each of the centroids after random noise was applied. This produces a valid discrete distribution, which crucially will typically assign the highest probability to the ground truth bin. One of the remaining discrete features was simply a copy of a previous feature, and the other was random noise with no relationship to the ground truth class.

A dataset consisting of 10,000 samples was generated according to these procedures, of which 7,000 were used for training models, and 3,000 were used for testing their resulting accuracies.

\subsection{Experimental Results}

\begin{figure}[h]
    \centering
    \includegraphics[width=\textwidth]{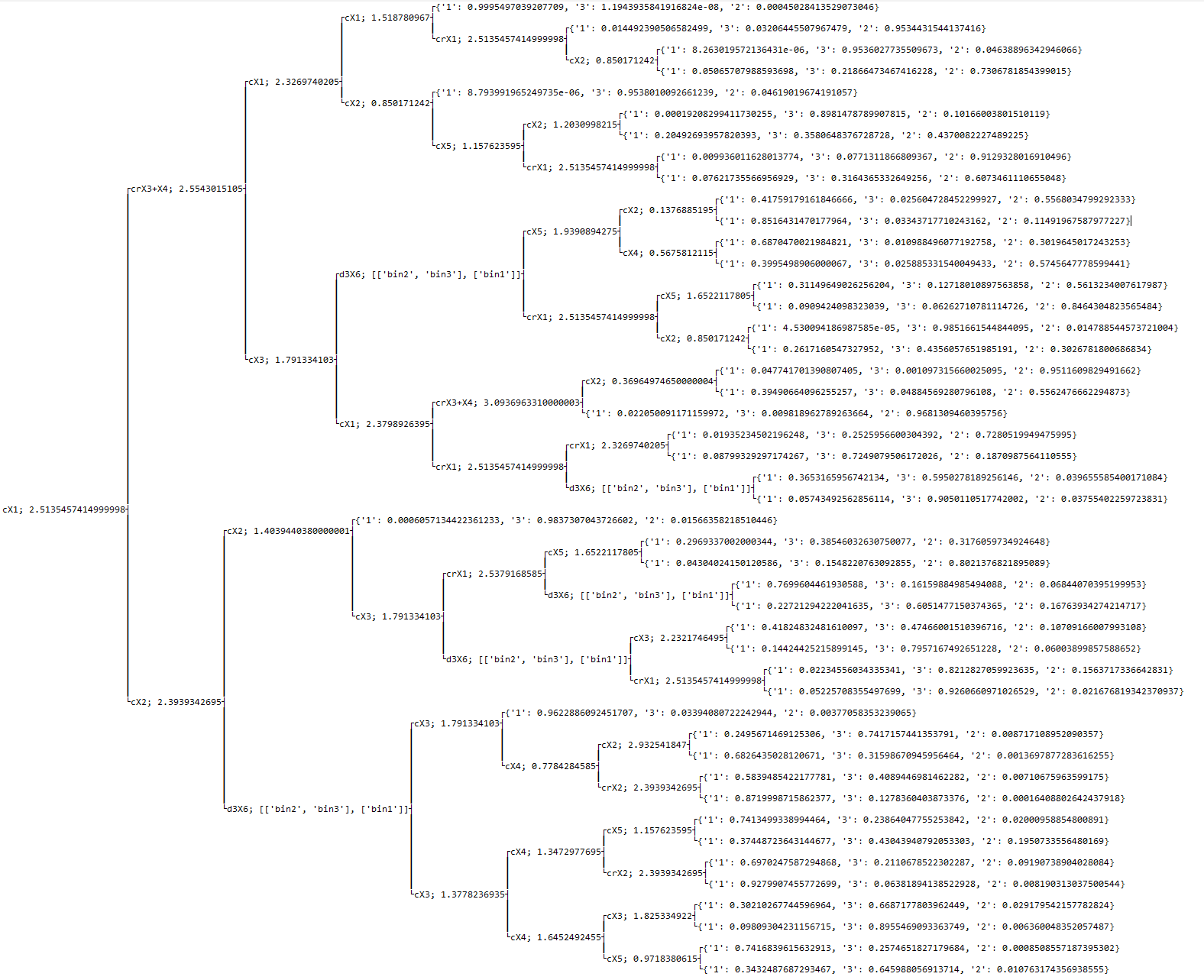}
    \caption{An Indecision Tree with a maximum depth of 6, trained on the synethic dataset, rendered with ASCII art.}
    \label{fig:ind_tree_ascii}
\end{figure}

The primary advance presented by this paper was the addition of handling quantified uncertainty to decision trees to produce indecision trees. Thus, the point of comparison to the indecision tree is to simply remove the uncertainty from the dataset, and run the same training and inference procedures on the same data. Here, the standard deviations were removed from the continuous features, and the bin probabilities were replaced with a 1 for the highest probability and a 0 for all others. Under these conditions, a standard decision tree achieved $90.6\%$ accuracy, as compared to the indecision tree's $98.1\%$ accuracy. This is a 7.5 point increase in accuracy, as well as a $79.8\%$ reduction in error. The probabilities provided to the model increased the amount of information that it had about the data, the distributions aided in the robustness of the inferences, and removing it lead to an expected and reasonable decrease in the accuracy of the model. Thus were the aims of developing this method validated, and we can now effectively leverage the quantified uncertainty of measurements to aid in learning and inference.

\section{Conclusions and Future Work}
As seen in the experimental results, providing the uncertainties to the model increased the amount of information that it had about the data, the distributions aiding in the robustness of the inferences, and removing it lead to an expected and reasonable decrease in the accuracy of the model. Thus were the aims of developing this method validated, and we can now effectively leverage the quantified uncertainty of measurements to aid in learning and inference.

Future and presently ongoing work will involve adapting other algorithms like MLNs and MLEM2 to work in this setting with quantified uncertainty, as well as trying to learn lessons for applications to other classification problems. Perhaps modifying neural networks to account for measurement uncertainty would similarly increase their robustness and accuracy, as well as harden them against adversarial methods.

\section*{Acknowledgements}

We would like to thank our coworker and colleague Hugh McLaughlin, for whom the descriptor ``tremendous help" would be woefully inadequate. Without his contributions as the only person involved who actually knows how to write software, this work doesn't get past the ``neat idea" stage.

\bibliography{main} 
\bibliographystyle{spiebib} 

\begin{center}
\scalebox{.2}{\textcolor{white}{``A little song, a little dance, a little seltzer down your pants." -Tim}}
\end{center}

\end{document}